\documentclass[10pt,twocolumn,letterpaper]{article}

\usepackage{cvpr}              %
\PassOptionsToPackage{square,numbers,sort}{natbib}
\usepackage{multirow} 
\usepackage{amsmath}
\usepackage{amsthm}
\usepackage{xcolor}
\usepackage{stfloats}
\usepackage[table]{xcolor}

\definecolor{cvprblue}{rgb}{0.21,0.49,0.74}
\definecolor{lightblue}{rgb}{0.8, 0.9, 1} 
\usepackage[pagebackref,breaklinks,colorlinks,allcolors=cvprblue]{hyperref}
\usepackage{subcaption}
\usepackage{graphicx,hyperref}

\def\paperID{} %
\def\confName{CVPR}
\def\confYear{2026}

\title{QVGGT: Post-Training Quantized Visual Geometry Grounded Transformer}

\author{
Zhizhen Pan$^{1, 2}$ \quad
Hesong Wang$^{1}$ \quad
Huan Wang$^{1, }$\thanks{Corresponding author.}\\
\textsuperscript{1}Westlake University\quad
\textsuperscript{2}Beijing University of Posts and Telecommunications\\
\texttt{\small{zhizhenpan9}@gmail.com}\\
\texttt{\small\{wanghesong, wanghuan\}@westlake.edu.cn}\\
{\small \url{https://ddsacu.github.io/QVGGT/}}
}

\begin{document}
\twocolumn[{
\maketitle
\begin{center}
\vspace{-10mm}
\includegraphics[width=0.96\linewidth]{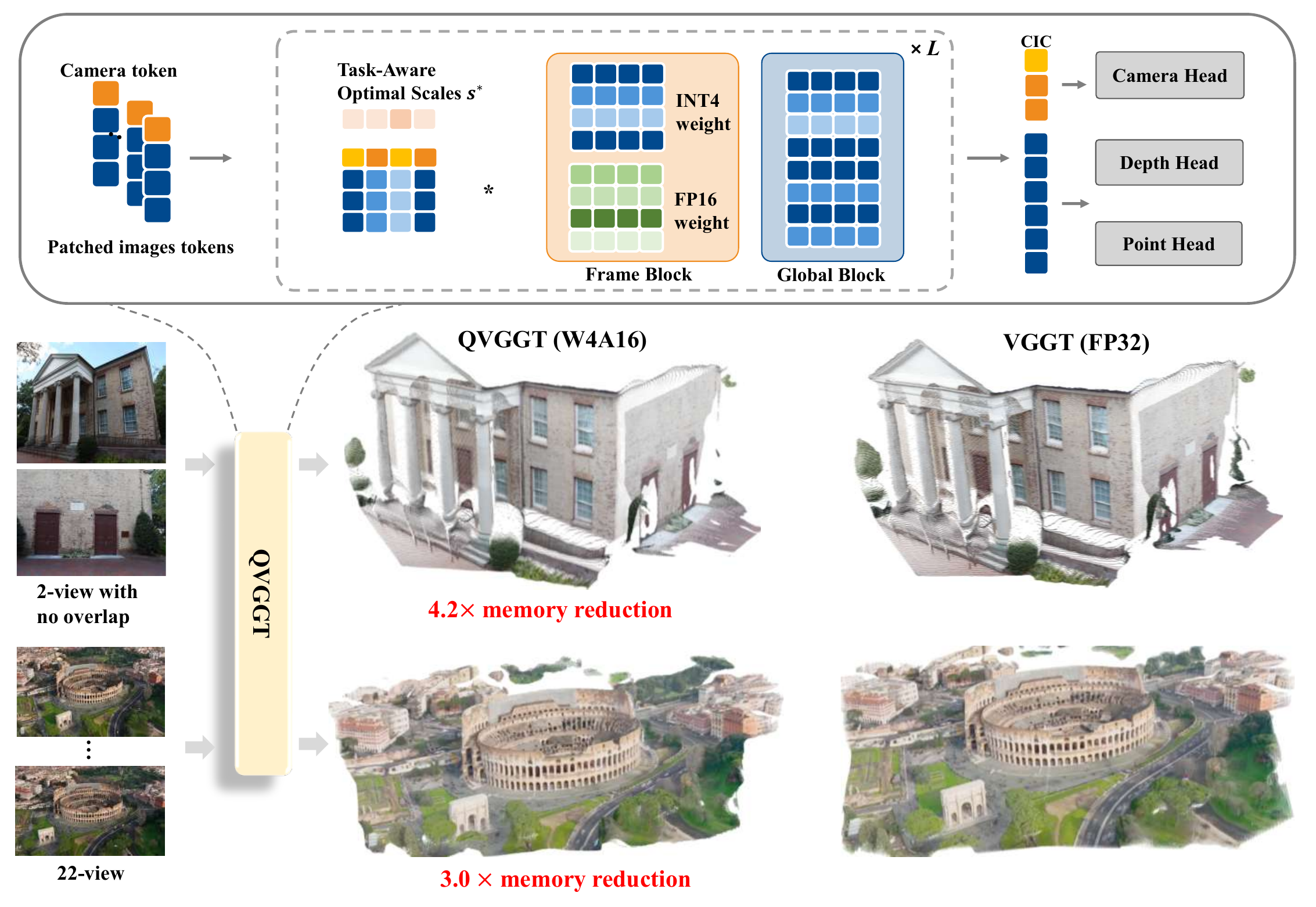}
\end{center}
\captionsetup{type=figure}
\vspace{-8mm}
\captionof{figure}{\textbf{Top:} {We introduce QVGGT, a three-stage post-training quantization framework that stabilizes VGGT by mixed-precision allocation, token filtering with PCA-based information compensation, and task-aware scale search preserving 3D geometric consistency. \textbf{Bottom:} Visualized 3D reconstruction results on real-world scenes show that QVGGT (W4A16) achieves comparable performance to VGGT (FP32), while delivering 4.2× and 3.0× memory reduction for  2 input images and 22 input images, respectively.}
}\label{fig:teaser}
\vspace{4mm} }]

\begin{NoHyper}
\def\thefootnote{*}
\footnotetext{Corresponding author.}

\def\thefootnote{$\dagger$}
\footnotetext{Work done as a visiting research intern at ENCODE Lab, Westlake University.}
\end{NoHyper}
\def\thefootnote{\arabic{footnote}}

\begin{abstract}
Estimating 3D attributes directly from images has advanced rapidly with the Visual Geometry Grounded Transformer (VGGT), which predicts camera parameters, depth maps, and point clouds in a single forward pass. However, its 1.2B-parameter scale severely limits deployment on resource-constrained platforms such as UAVs and mobile AR devices. To address this limitation, we introduce QVGGT, a tailored quantization framework designed to compress VGGT. 
Our approach starts from the observation that transformer blocks within VGGT exhibit heterogeneous sensitivity to quantization. We thus analyze per-block quantization sensitivity and propose a selective mixed-precision strategy that allocates higher precision to the most fragile transformer blocks. To address the amplification of quantization error caused by high-variance camera and register tokens, we further introduce token filtering with camera information compensation, which removes these outliers from activation calibration and restores their geometric cues using a PCA-derived global compensation token. Finally, we develop a task-aware scale search mechanism that evaluates candidate quantization scales not only through layer reconstruction but also through multi-head supervision and cross-head geometric consistency among camera poses, depth maps, and point maps.
Extensive experiments on multiple geometry perception benchmarks demonstrate that QVGGT achieves near-lossless W4A16 quantization, preserving the accuracy of all 3D prediction heads while delivering 3$\sim$4.9$\times$ memory reduction and up to 2.8$\times$ real hardware speedup over FP32.
Our approach makes high-fidelity 3D perception feasible on edge devices, enabling practical deployment of feed-forward 3D reconstruction models in real-world constrained environments. 
\end{abstract}
    
\section{Introduction}
\label{sec:intro}

Estimating the 3D attributes of a scene from visual inputs is a cornerstone of computer vision, with broad applications in robotics, augmented reality, and digital content creation. Traditional 3D reconstruction methods such as Structure-from-Motion (SfM) \citep{liu2025robust, Wang_2024_VGGSfM} and Multi-View Stereo (MVS) \citep{gu2020cascade, ma2022multiview} rely on explicit geometric constraints such as Bundle Adjustment (BA) \citep{hartley_multiple_2000}, which, while accurate, involve iterative optimization and high computational cost. Recent advances in learning-based approaches, such as DUSt3R \citep{wang2024dust3r} and MASt3R \citep{leroy2024mast3r}, have demonstrated that feed-forward neural networks can directly predict 3D attributes, significantly simplifying the reconstruction pipeline. Building on this line of progress, the Visual Geometry Grounded Transformer (VGGT) \citep{wang2025vggt} has emerged as a breakthrough model, capable of predicting all key 3D attributes in a single forward pass.

Despite its effectiveness, VGGT contains 1.26 billion parameters, resulting in a substantial memory footprint and computational demand during inference. This makes its deployment challenging in real-time and resource-constrained scenarios such as robotic navigation and AR/VR perception, where both efficiency and geometric accuracy are crucial. Among common model compression techniques, model pruning \citep{han2016deep, zhang2023advancingmodelpruningbilevel, sun2023simple, frantar2022obc, zhu2025obs} and knowledge distillation \citep{hinton2015distilling, zhao2022decoupledknowledgedistillation, NEURIPS2022_huan, NEURIPS2021_huan} reduce model complexity but offer limited acceleration on modern hardware. In contrast, quantization \citep{ashkboos2024quarot, dettmers2022llmint88bitmatrixmultiplication, xiao2023smoothquant, NEURIPS2022_Zeroquant, liu2024spinquant} simultaneously improves compactness and speed by reducing numerical precision. Although quantization has enabled efficient on-device inference for large language models \citep{lin2024awq, zhao2024atom}, quantizing large transformer-based 3D reconstruction networks remains largely unexplored.

We propose QVGGT, a post-training quantization framework tailored for 3D reconstruction transformers. 
First, we observe that different blocks in VGGT exhibit heterogeneous sensitivity to quantization, which makes uniform low-bit quantization suboptimal. To address this, we perform a layer-wise sensitivity analysis and adopt a selective mixed-precision strategy that preserves higher precision for sensitive blocks while quantizing the remaining layers to 4 bits.
Second, we identify that camera and register tokens produce high-magnitude activations that can dominate scale estimation during calibration, amplifying quantization error. To mitigate this issue, we introduce a token filtering mechanism that excludes these tokens from activation statistics, together with a camera information compensation (CIC) strategy that restores their influence at inference time via a PCA-based token.
Finally, we note that standard layer-wise reconstruction objectives do not fully reflect downstream geometric performance. Therefore, we propose a task-aware scale search mechanism, where quantization scales are selected based on multi-head supervision and a geometric consistency objective, aligning scale optimization with 3D reconstruction quality.

We evaluate QVGGT on a suite of geometry perception benchmarks, including CO3Dv2~\citep{reizenstein2021co3d}, RealEstate10K~\citep{RealEstate10K}, 7-Scenes~\citep{shotton2013scene}, and NRGBD~\citep{azinovic2022neural}. 
Under a practical W4A16 setting, QVGGT yields near-lossless camera pose estimation and reduces model size by over 75\%. 
Compared with general-purpose PTQ methods such as SmoothQuant~\citep{xiao2023smoothquant}, GPTQ~\citep{frantar2022gptq}, and AWQ~\citep{lin2024awq}, QVGGT consistently preserves higher accuracy, highlighting the difficulty of directly applying standard quantization techniques to VGGT. 
Furthermore, under matched quantization settings, QVGGT achieves comparable or better performance than the concurrent QuantVGGT, while maintaining favorable efficiency. 

To summarize, we make the following contributions:
\begin{itemize}
    \item We present a PTQ framework for 3D geometry transformers.  We observe the heterogeneous quantization sensitivity of VGGT and show that it can be effectively leveraged via a selective mixed-precision strategy.
    
    \item We identify a characteristic token-level activation imbalance in 3D transformers, and propose a token-filtered calibration scheme with compensation to improve the robustness of activation-based quantization.
    
    \item We design a task-aware scale search mechanism with multi-head supervision and geometric consistency, aligning quantization with downstream 3D prediction tasks.
    
    \item Empirically, QVGGT achieves nearly lossless W4A16 quantization across multiple 3D geometry prediction benchmarks, while reducing model size by more than 75\% and decreasing the inference memory by 3$\sim$4.9×.
\end{itemize}

\begin{figure*}[t]
    \vspace{-10mm}
    \centering
    \includegraphics[width=0.98\textwidth]{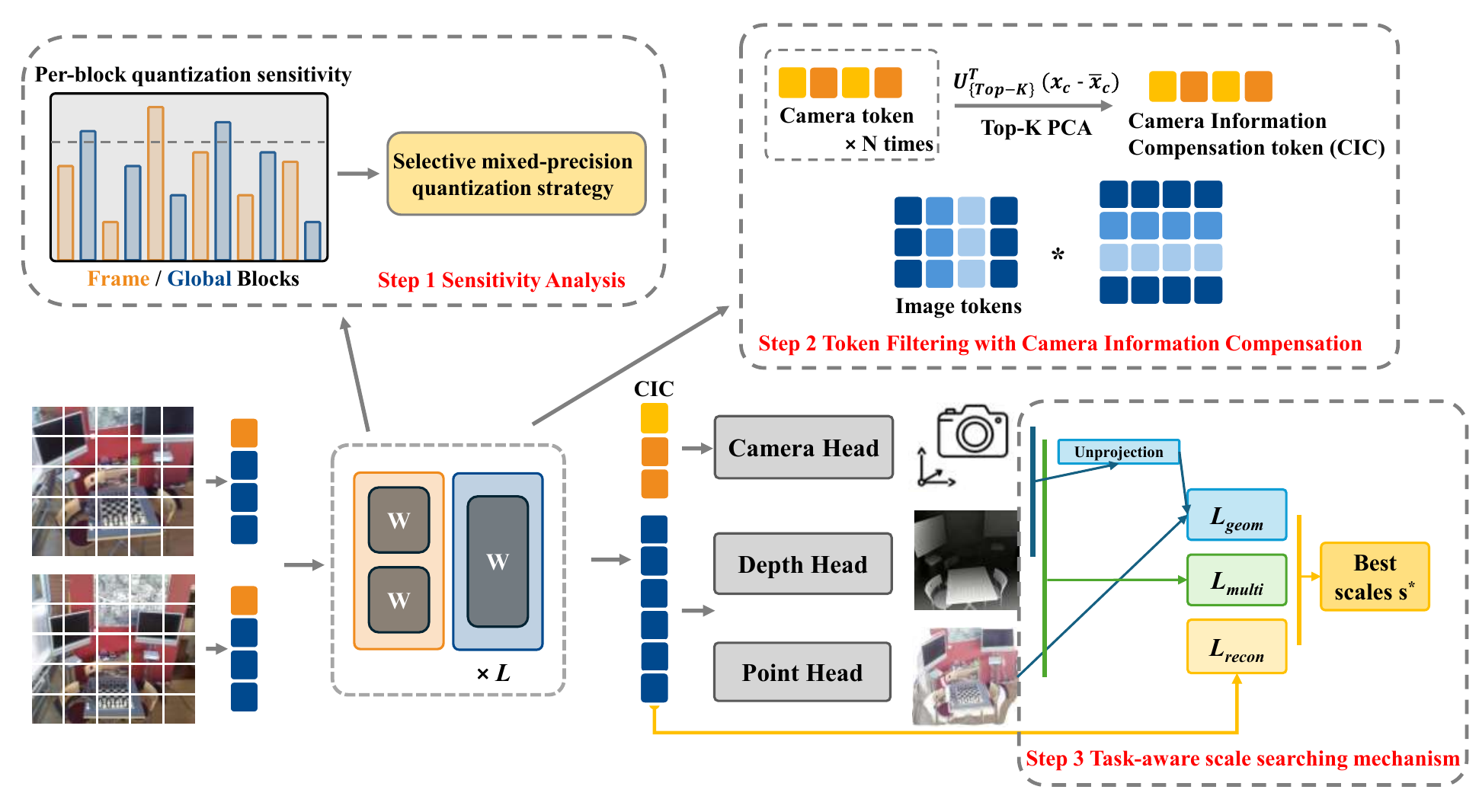}
    \vspace{-5mm}
    \caption{
    \textbf{Overview of QVGGT.} 
    Our framework consists of three components to preserve VGGT performance under low-bit quantization. 
    \textbf{Step 1: Sensitivity analysis.} We estimate per-block quantization sensitivity across frame-wise and global transformer blocks, enabling selective mixed-precision assignment for critical layers. 
    \textbf{Step 2: Token filtering with camera information compensation.} To mitigate outlier-dominated scale estimation, we exclude camera and register tokens during activation statistics collection. A camera information compensation (CIC) token is then constructed via top-$K$ PCA over camera-token activations and injected into the camera head. 
    \textbf{Step 3: Task-aware scale search.} We select quantization scales using a task-aware objective that combines multi-head losses, geometric consistency, and reconstruction error, jointly optimizing accuracy and robustness across all heads.
    }
    \vspace{-4mm}
    
\label{fig:workflow}
\end{figure*}

\section{Related Work}
\label{sec:related work}

\textbf{Learning-based 3D Reconstruction.}

\noindent Over the past decade, estimating 3D attributes including camera pose, depth, and point maps has been a core research problem in 3D computer vision \cite{hartley2003multiple, galliani2015massively}.
Recent advances in deep learning have significantly shifted 3D reconstruction from traditional geometry-based pipelines~\citep{agarwal2011building, frahm2010building, liu2025robust, schonberger2016structure, schonberger2016pixelwise, Wei_2021_NerfingMVS} toward data-driven approaches. Learning-based methods have been proposed for exploring end-to-end 3D reconstruction frameworks~\citep{yang2024depth, brachmann2024acezero, yang2025fast3r, tang2024mv, Wang_2024_VGGSfM, ma2022multiview, wang2025cut3r}, which have demonstrated the feasibility of directly predicting dense point maps from image pairs, thereby reducing reliance on explicit post-processing.
Representative works such as DUSt3R~\citep{wang2024dust3r} formulate 3D reconstruction as direct regression of point maps from image pairs, while MASt3R~\citep{leroy2024mast3r} further improves metric consistency through confidence-aware optimization. More recently, VGGT~\citep{wang2025vggt} unifies multiple 3D perception tasks within a single transformer-based architecture.
However, the large model size and high computational cost of VGGT significantly hinder its practical deployment. In particular, compression techniques for large-scale 3D transformers, such as post-training quantization, remain largely underexplored.

\noindent\textbf{Model Quantization.}

\noindent Model quantization reduces memory and computational overhead by mapping floating-point parameters to low-bit integers, primarily through Quantization-Aware Training (QAT) or Post-Training Quantization (PTQ). QAT \citep{jacob2018quantizationandtrain, liu2024llmqat, chen-etal-2025-efficientqat} achieves high accuracy at extreme bit-widths by simulating quantization during training, yet its intensive resource demand makes it impractical for billion-parameter models like VGGT. Consequently, PTQ has emerged as the preferred paradigm for Large Language Models (LLMs) \citep{dettmers2022llmint88bitmatrixmultiplication, frantar2022gptq, xiao2023smoothquant, ashkboos2024quarot, zhao2024atom, lin2024awq}, Vision Transformers (ViTs)~\citep{yang2024dopqvit, yuan2022ptq4vit}, and MLLMs ~\citep{tao2025plug, yu2025mquant}. The field has progressed from outlier-aware mixed-precision \citep{dettmers2022llmint88bitmatrixmultiplication} and second-order weight compensation \citep{frantar2022gptq} toward activation-weight co-scaling \citep{xiao2023smoothquant, lin2024awq} and mathematical distribution reshaping via rotational transforms \citep{ashkboos2024quarot} or fine-grained modeling \citep{yuan2022ptq4vit, yang2024dopqvit}. 

\enlargethispage{0.5\baselineskip}

While these approaches effectively stabilize numerical distributions of language and 2D vision models, they generalize poorly to 3D geometry networks such as VGGT~\citep{wang2025vggt}. 
A concurrent work, QuantVGGT~\citep{feng2025quantized}, explores post-training quantization for VGGT under specific bit-width settings and demonstrates competitive performance. 
In contrast, our approach differs in both quantization settings and methodology design, aiming to better accommodate the unique characteristics of 3D geometry transformers. We defer detailed comparisons to the end of the method section.

\vspace{-2mm}

\section{Method}
\label{headings}

\begin{figure*}[t]
\centering
\resizebox{0.998\linewidth}{!}{  %
\setlength{\tabcolsep}{1mm}  %
\renewcommand{\arraystretch}{0.8}  %
\begin{tabular}{@{}ccc}  %
\begin{subfigure}{0.38\linewidth}
  \includegraphics[width=\linewidth]{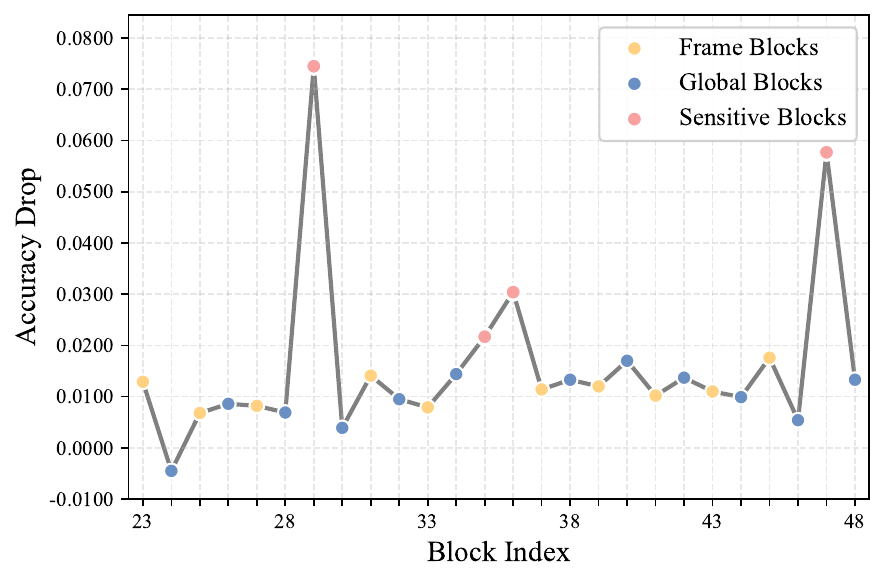}
  \subcaption{}
  \label{fig: sensitivity analysis}
\end{subfigure} &
\begin{subfigure}{0.32\linewidth}
  \includegraphics[width=\linewidth]{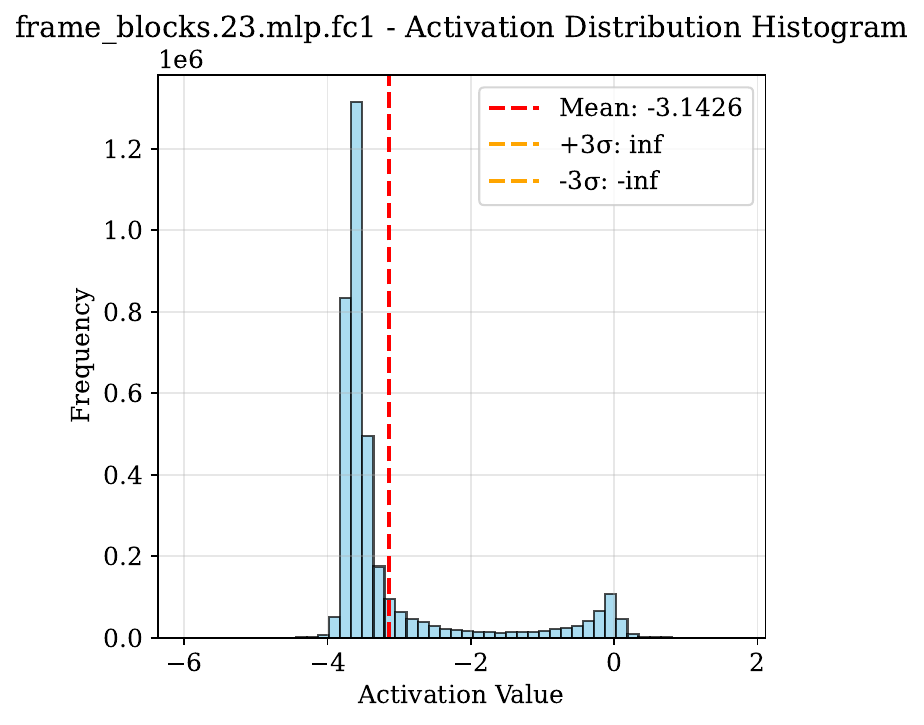}
  \subcaption{}
  \label{fig: outlier histogram}
\end{subfigure} &
\begin{subfigure}{0.26\linewidth}
  \includegraphics[width=\linewidth]{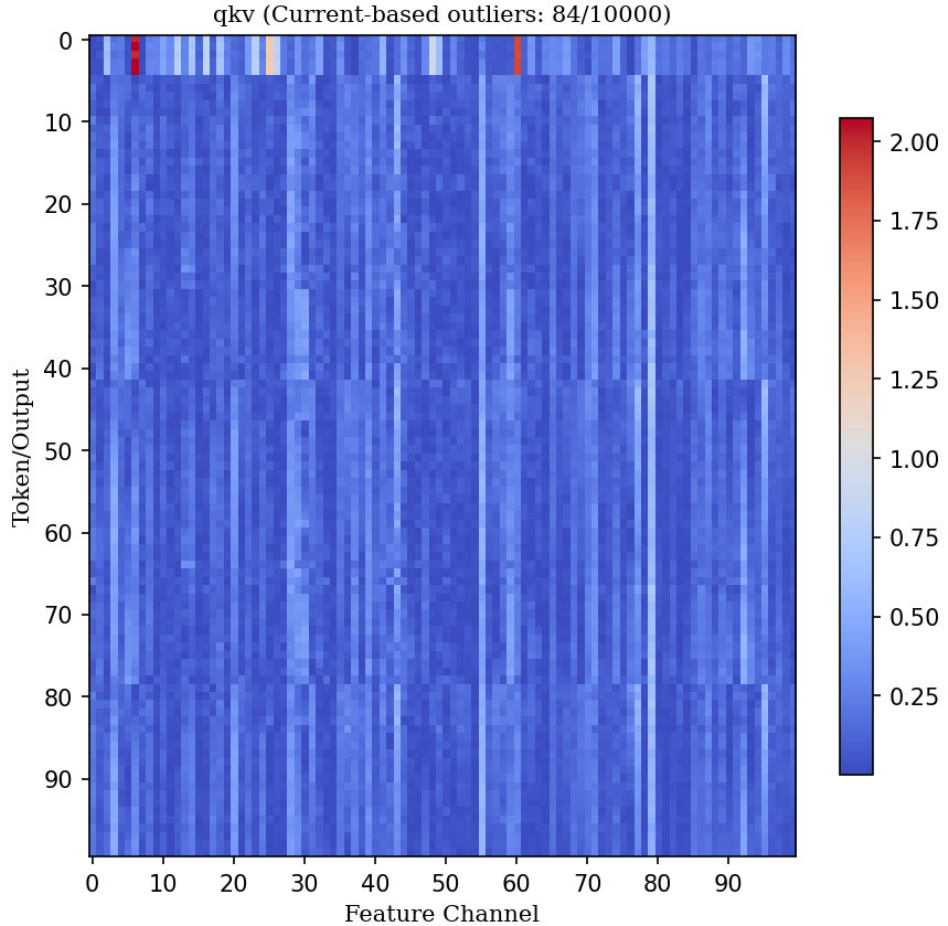}
  \subcaption{}
  \label{fig: outlier heatmap}
  \end{subfigure}
\end{tabular}}
\vspace{-4mm}
\caption{\textbf{Motivation of selective mixed-precision quantization:} (a) Per-block sensitivity analysis of VGGT. The abscissa shows the alternating arrangement of frame blocks and global blocks; the ordinate represents the accuracy drop of camera pose prediction on the CO3Dv2 dataset when each block is quantized individually. (b) The outlier distribution histogram of frame block 23 shows that some sensitive blocks have a long-tail distribution, which will lead to the amplification of quantization errors. (c) Activation outlier distribution of frame block 23, where the camera and register tokens exhibit distinct high-variance activations compared to patch tokens, leading to amplified quantization errors in per-channel schemes.}
\vspace{-4mm}
\label{fig:activation_distribution}
\end{figure*}
We first introduce the preliminaries of QVGGT (Section ~\ref{3.1 prelim}). We perform the selective mixed-precision quantization (Section \ref{3.2 selec}). Then we smooth the camera token to reduce quantization error (Section \ref{3.3 CIC}). Finally, we design a task-aware scale search methodology (Section \ref{3.4 task-aware}).

\subsection{Preliminary}
\label{3.1 prelim}

\textbf{Visual Geometry Grounded Transformer} (VGGT) is a feed-forward architecture designed to estimate comprehensive 3D scene attributes directly from multi-view RGB inputs. Given a sequence of $N$ RGB images $(I_i)_{i=1}^N$, where each image $I_i \in \mathbb{R}^{3 \times H \times W}$ observes the same static 3D scene, VGGT maps this sequence to a corresponding set of geometric outputs:
\begin{equation}
f_\theta((I_i)_{i=1}^N) = \{(g_i, D_i, P_i)\}_{i=1}^N,
\end{equation}
where each component is defined as follows: $g_i \in \mathbb{R}^9$ denotes camera parameters consisting of a rotation quaternion $q_i \in \mathbb{R}^4$, translation vector $t_i \in \mathbb{R}^3$, and field-of-view $f_i \in \mathbb{R}^2$; $D_i \in \mathbb{R}^{H \times W}$ represents the depth map; and $P_i \in \mathbb{R}^{3 \times H \times W}$is the point map in 3D world coordinates.

Input images $I_i$ are patched into a set of tokens $t_I \in \mathbb{R}^{K \times C}$ through DINO \citep{oquab2023dinov2}. A camera token $t^C_i \in R^{1\times C}$ and four register tokens $t^i_R \in R^{4\times C}$ are appended to form the complete token sequence for each frame. 
The resulting token sequences are then processed by the VGGT backbone, an alternating-attention (AA) transformer composed of interleaved frame blocks and global blocks, where the former captures intra-frame features while the latter aggregates cross-frame features. Finally, the output tokens are passed to task-specific prediction heads, including the camera head and the DPT \citep{ranftl21dpt} head (for depth map and point map prediction), to produce the final 3D attribute.

\vspace{0.5em}
\noindent\textbf{Post Training Quantization} (PTQ) reduces model size and inference cost by mapping floating-point parameters to low-bit integers without retraining. Unlike quantization-aware training (QAT), which updates both weights and quantization parameters via backpropagation, PTQ quantizes a pretrained model using a calibration dataset.

In this work, we focus on weight-only PTQ, where the activation remains in floating-point while the model weights are quantized to lower precision.

For a weight element $w \in W$, the uniform quantization functions are defined as: 
\begin{equation}
Q(w) = \Delta \cdot \text{Round}\left( \frac{w}{\Delta} \right), \quad \Delta = \frac{\max(|w|)}{2^{N - 1}}, 
\end{equation}
where $x$ is the input activation, $W$ is the weight, $Q(W)$ is its quantized counterpart, $N$ is the bit-width, and $\Delta$ determines the quantization step size based on the maximum absolute value in the group (or channel).

After quantization, the dequantized weight $\hat{w}$ can be expressed as: 
\begin{equation}
\hat{w} = Q(w) = \Delta \cdot q, \quad q \in \{-2^{N-1}, \dots, 2^{N-1} - 1\}.
\end{equation}
This symmetric scheme effectively reduces the storage cost while maintaining arithmetic compatibility with existing GPU and accelerator instructions.

\subsection{Selective mixed-precision quantization}
\label{3.2 selec}

\textbf{Heterogeneous Quantization Sensitivity of VGGT Blocks.} We start our study by observing that directly applying standard quantization to all Alternating Attention (AA) blocks of VGGT leads to prominent and highly uneven accuracy degradation. To gain deeper insights into how different parts of VGGT respond to low-bit quantization, we conduct a systematic and fine-grained sensitivity analysis by selectively quantizing individual blocks and measuring the resulting impact on downstream prediction heads.

Interestingly, the camera head exhibits the strongest and most volatile dependence on block precision. As shown in Figure \ref{fig: sensitivity analysis}, quantizing certain blocks introduces disproportionately large errors in camera pose estimation, even though the depth and point-map heads remain comparatively stable and robust. This analysis reveals a highly heterogeneous and task-dependent sensitivity pattern within AA blocks: while some blocks can be safely quantized with negligible performance loss, others trigger a dramatic collapse in accuracy.
These findings highlight that camera pose estimation serves as a particularly sensitive and informative indicator for quantization stability.

Using the same methodology, we conducted a sensitivity analysis on the different linear layers within the AA Blocks. We find that the attention projection layer and the first FFN layer exhibit the lowest and highest quantization sensitivity.

Our sensitivity analysis (Figure \ref{fig: sensitivity analysis}) reveals that different AA blocks exhibit heterogeneous robustness to quantization. To maximize model compression and minimize quantization errors, we adopt a selective mixed-precision quantization strategy. Based on the sensitivity analysis, the 14th, 17th, and 23rd frame blocks and the 23rd global block are sensitive to quantization, for which we have chosen to only quantize the attention projection layer while keeping the weights of the remaining layers at FP16 precision. The other blocks exhibit relatively high quantization robustness, so we perform INT4 quantization on them. 

\subsection{Camera Information Compensation}
\label{3.3 CIC}

\textbf{Outlier distribution of camera token and register tokens.} Section \ref{3.2 selec} reveals the camera head is disproportionately vulnerable to block-wise quantization. To investigate the underlying cause, we visualize the activation statistics of each transformer block and identify that camera tokens (the sole inputs to the camera head) and register tokens exhibit \textit{outlier activation magnitudes} in several quantization-sensitive layers. 
As shown in Figure~\ref{fig: outlier histogram} and Figure~\ref{fig: outlier heatmap}, the 2D activation heatmap of the 23rd frame block reveals a pronounced deviation: camera and register tokens consistently lie in high-magnitude regions of the activation distribution, while image tokens form a compact cluster around the mean. 
We empirically observe that these outlier activations are strongly correlated with camera pose estimation performance, as perturbing them leads to noticeable degradation in the camera head accuracy. This suggests that these tokens carry information that is particularly sensitive for downstream geometric prediction. 

Such heterogeneity in token statistics directly affects activation-aware weight quantization. Per-channel (or per-group) weight scales $s$ are optimized according to activation distribution, with the objective of minimizing the error:
\begin{equation}
\begin{gathered}
s^* = \arg\min_{s} \mathcal{L}(s), \\
\mathcal{L}(s) =
\left\lVert
Q\!\left(W \operatorname{diag}(s)\right)
\operatorname{diag}(s)^{-1} X
- WX
\right\rVert,
\end{gathered}
\end{equation}
where \( X = [t^i_C; t^i_R; t^i_I] \) is the concatenation of camera, register, and image tokens. When camera and register tokens exhibit significantly larger magnitudes, the optimization becomes dominated by these high-amplitude tokens, biasing the scale estimation toward a small subset of activations and degrading overall quantization fidelity.

These large-magnitude outliers inflate the dynamic range of the corresponding activation channels. Consequently, the quantization scale is dominated by a few extreme activations, leading to an unbalanced quantization resolution—fine-grained channels for camera-related weights but coarse quantization for others—thereby amplifying the overall quantization error.

\vspace{0.5em}
\noindent\textbf{Camera-token-filtered quantization strategy.} 
To address this bias, we propose a camera-token-filtered quantization strategy that disentangles scale estimation from high-variance tokens while preserving their camera information during inference.
Specifically, during the calibration phase, we compute activation statistics and perform scale search only using image tokens. This ensures that the optimized \(\quad s^*\) reflects the dynamic range of the majority token population rather than being pulled by a few extreme activations.

\vspace{0.5em}
\noindent\textbf{Camera information compensation.}
Excluding camera tokens during scale search stabilizes quantization but removes the influence of dense global cues essential for the camera head. To compensate for this information loss, we synthesize a camera information compensation token (CICT) derived from the camera-token distribution observed on the calibration set.
This CICT encapsulates low-variance, dataset-level global geometry priors that are later re-injected into the camera head at inference.

\noindent Let the calibration camera-token matrix be: 
\begin{equation}
X_c = \left[ x_c^{(1)\top}; \, x_c^{(2)\top}; \, \ldots; \, x_c^{(N)\top} \right] \in \mathbb{R}^{N \times 2C}.
\end{equation}
Center $X_c$ and compute its top-$K$ principal directions $U_K \in \mathbb{R}^{D \times K}$. Project each sample: 
\begin{equation}
z^{(i)} = U_K^\top \left( x_c^{(i)} - \mu \right), \quad \mu = \frac{1}{N} \sum_i x_c^{(i)}.
\end{equation}
Form the mean projection $\bar{z} = \frac{1}{N} \sum_i z^{(i)}$ and reconstruct the camera information compensation token: 
\begin{equation}
\tilde{x}_{\text{CICT}} = \mu + U_K \bar{z}.
\end{equation}
Normalize scale to match patch-token norms (so attention magnitudes remain compatible): 
\begin{equation}
x_{\text{CICT}} = \tilde{x}_{\text{CICT}} \cdot \frac{\overline{\| x_p \|_2}}{\| \tilde{x}_{\text{CICT}} \|_2},
\end{equation}
where $\overline{\| x_p \|_2}$ is the mean $L_2$ norm of representative patch tokens from calibration.         
During inference, after the model generates the per-frame token sequence $\{ x_c^{(t)} \}_{t=1}^S$, the synthesized $x_{\text{CICT}}$ is appended to the token sequence before it is passed to the camera head.

\subsection{Task-aware Scale Searching Mechanism}
\label{3.4 task-aware}

Activation-aware quantization determines the per-channel (or per-group) scaling factors $s$ by minimizing the output reconstruction error after quantization.

Although effective for preserving layer-wise numerical fidelity, this objective does not reflect the impact of quantization on task-level predictions.
In VGGT, where camera pose, depth, and point maps interact through strict geometric relationships, layer reconstruction quality does not necessarily guarantee accurate 3D attribute estimation.
To better align quantization with task objectives, we introduce a task-aware scale search mechanism that supervises scale selection using both multi-head prediction losses and geometric consistency constraints.

\vspace{2mm}
\noindent\textbf{Multi-head loss.} Instead of minimizing only the layer reconstruction loss, we incorporate task-level supervision derived from VGGT's three prediction heads. Specifically, given the predicted camera parameters \( \hat{g}_i \), depth maps \( \hat{D}_i \), and point maps \( \hat{P}_i \), with corresponding ground truths \( g_i \), \( D_i \), \( P_i \), we define three head-specific task losses: 
\begin{equation}
\begin{gathered}
L_{\text{camera}} = \sum_{i=1}^{N} \|\hat{g}_i - g_i\|_{\varepsilon}, \\
L_{\text{depth}} = \sum_{i=1}^{N} \|\hat{\Sigma}_i^D \odot (\hat{D}_i - D_i)\|, \\
L_{\text{point}} = \sum_{i=1}^{N} \|\hat{\Sigma}_i^P \odot (\hat{P}_i - P_i)\|,
\end{gathered}
\end{equation}
where $\|\cdot\|_{\varepsilon}$ denotes the Huber loss, and $\odot$ is the element-wise product weighted by the predicted uncertainty maps.

\vspace{2mm}
\noindent\textbf{Geometry consistency loss.} VGGT is trained in a multi-head fashion, where camera, depth, and point predictions are mutually dependent via geometric constraints. While these heads are designed to capture different geometric attributes, they are inherently connected due to geometric consistency. For example, the combination of predicted camera pose and depth can reconstruct a point map, which should agree with the direct output of the point-map head. Although this redundancy was originally introduced to improve training accuracy, we leverage it in our quantization framework by introducing a geometry consistency loss.

Formally, given the predicted depth map \( \mathbf{D} \), camera intrinsics \( \mathbf{K} \), and extrinsics \( \mathbf{E} = (\mathbf{R}, \mathbf{t}) \), we unproject depth values into 3D space:
\begin{equation}
\mathbf{W}^{\text{proj}}(u, v) = \mathbf{R}^{-1} \left( \frac{D(u, v)}{f} \begin{bmatrix} u - c_x \\ v - c_y \\ f \end{bmatrix} - \mathbf{t} \right),
\end{equation}
where \( (u, v) \) denotes the image pixel, \( f \) the focal length, and \( (c_x, c_y) \) the principal point. This yields a reconstructed point map \( \mathbf{W}^{\text{proj}} \). 

Meanwhile, the point head directly outputs a dense point map $\mathbf{W}^{direct}$. The geometry consistency loss is defined as: 
\begin{equation}
\small\mathcal{L}_{\text{geom}} = \frac{1}{|\Omega|} \sum_{(u, v) \in \Omega} \left\| \mathbf{W}^{\text{direct}}(u, v) - \mathbf{W}^{\text{proj}}(u, v) \right\|_2,
\end{equation}
where $\Omega$ denotes the set of valid pixels.
This loss enforces cross-head agreement during scale search, penalizing discrepancies between reconstructed and predicted geometry, thereby guiding the selection of quantization scales toward solutions that preserve the intrinsic 3D structure across all heads.
The final task-aware objective for scale selection thus becomes: 
\begin{equation}
L_{\text{task}}(s) = L_{\text{recon}}+L_{\text{camera}} + \alpha L_{\text{depth}} + \beta L_{\text{point}} + L_{\text{geo}},
\end{equation}
where $\alpha$ and $\beta$ balance the relative importance of depth and point-map supervision. 
Since the camera, depth, and point-map losses exhibit comparable numerical ranges in practice, we assign equal weights by setting \( \alpha=\beta=1 \). Following the lightweight scale-search paradigm of AWQ~\citep{lin2024awq}, we perform grid search over candidate quantization scales, while replacing the reconstruction-based objective with the proposed task-aware optimization.

\vspace{2mm}
\noindent\textbf{Distinction from QuantVGGT.} 
We note a concurrent work, QuantVGGT~\cite{feng2025quantized}, which also studies PTQ for VGGT. While QuantVGGT primarily employs general-purpose numerical stabilization techniques (e.g., outlier dispersion and activation smoothing), QVGGT is designed from a geometry-aware perspective, tailored specifically to 3D transformers. 
In particular, instead of treating camera tokens purely as outliers, we explicitly account for their impact on camera prediction through the proposed Camera Information Compensation (CIC). In addition, our scale search is guided by a task-aware objective that incorporates multi-head geometric consistency. Overall, QVGGT emphasizes task- and structure-aware quantization, complementing the distribution-oriented design of QuantVGGT.

\section{Experiments}
\label{others}

\subsection{Experimental Setups}
\noindent \textbf{Models and benchmarks.} We adopt VGGT-1B~\citep{wang2025vggt} as our base model for all experiments. For the quantization configuration, we choose the W4A16 setting, which has better hardware adaptability and has been widely shown to offer an optimal balance between efficiency and accuracy \cite{lin2024awq}. We further apply per-group weight quantization, which has been shown to better retain model performance compared with per-channel or per-tensor quantization in both language and vision transformers~\citep{dettmers2023case}.
We evaluate QVGGT on four representative 3D reconstruction and geometry perception benchmarks: CO3Dv2~\citep{reizenstein2021co3d}, RealEstate10K~\citep{RealEstate10K}, 7-Scenes~\citep{shotton2013scene}, and NRGBD \citep{azinovic2022neural}. These datasets collectively cover diverse scenarios, such as object-level and scene-level, real-world data and synthetic data. We evaluate QVGGT on two tasks: camera pose estimation and 3D reconstruction performance, and assess both geometric precision and consistency after quantization.

\vspace{2mm}
\noindent\textbf{Implementation details}. All experiments are conducted on an RTX 4090 (24GB) with PyTorch~\citep{pytorch}. Note that the proposed quantization method is post-training, namely, with no training cost. We adapt the HuggingFace Transformers~\citep{wolf2019huggingface} library to re-implement VGGT~\citep{wang2025vggt} as our baseline model.

\vspace{2mm}

\noindent\textbf{Comparison methods.} 
We compare QVGGT against several state-of-the-art PTQ methods originally designed for vision and language models, including SmoothQuant~\citep{xiao2023smoothquant}, GPTQ~\citep{frantar2022gptq}, and AWQ~\citep{lin2024awq}, as well as QuantVGGT~\citep{feng2025quantized}, a concurrent work tailored for VGGT.

For generic PTQ baselines, we follow their standard configurations (e.g., W8A8 for SmoothQuant) and additionally evaluate them under matched settings (e.g., W4A16) when applicable to ensure fair comparison. 
For QuantVGGT, we report both its original results and matched-bitwidth comparisons when available. 
We also report the performance of the full-precision VGGT as the upper bound.

\subsection{Camera Pose Estimation }

\begin{table}[t]
\footnotesize
\centering
\caption{Camera pose estimation on CO3Dv2~\citep{reizenstein2021co3d} and RealEstate10K~\citep{RealEstate10K} with 10 random frames.}
\vspace{-1mm}
\label{tab:co3dv2_results}
\scalebox{0.9}{
\begin{tabular}{l c@{\hspace{3pt}} c@{\hspace{4pt}}c c@{\hspace{4pt}}c}
    \toprule
    \multirow{2}{*}{\textbf{Method}} & \multirow{2}{*}{\textbf{W/A}} 
    & \multicolumn{2}{c}{\textbf{CO3Dv2}} & \multicolumn{2}{c}{\textbf{Re10K}} \\
    \cmidrule(lr){3-4} \cmidrule(lr){5-6}
    & & AUC@30$\uparrow$ & Latency 
    & AUC@30$\uparrow$ & Latency \\
    \midrule
    \rowcolor{gray!20}
    Baseline & FP16 & 89.5 & 0.38s & 85.3 & 0.37s \\
    \midrule
    SmoothQuant \citep{xiao2023smoothquant} & W8A8 & 87.9 & 0.62s & 81.3 & 0.38s\\
    QuantVGGT \citep{feng2025quantized} & W8A8 & 89.4 & - & 84.9 & - \\
    \midrule
    QuantVGGT~\citep{feng2025quantized} & W4A16 & 89.2 & - & 84.4 & - \\
    GPTQ \citep{frantar2022gptq} & W4A16 & 76.9 & 0.28s & 75.6 & 0.27s  \\
    AWQ \citep{lin2024awq} & W4A16 & 54.6 & 0.28s & 59.2 & 0.28s  \\
    \rowcolor{lightblue}
    QVGGT (ours) & W4A16 & \textbf{89.4 }& \textbf{0.23s} & \textbf{85.0} & \textbf{0.23s} \\
    \bottomrule
\end{tabular}
}
\vspace{-4mm}
\end{table}

\noindent We first evaluate our method on the CO3Dv2~\citep{reizenstein2021co3d} and RealEstate10K~\citep{RealEstate10K} datasets for camera pose estimation, as shown in Table~\ref{tab:co3dv2_results}. For each scene, we randomly sample 10 images and report AUC@30, which jointly measures relative rotation and translation accuracy.

To ensure a fair comparison, we include results under matched quantization settings (W4A16). Under this setting, QVGGT achieves constantly better performance across both datasets, while consistently delivering lower inference latency. 
It should be noted that the calibration data is sampled from Co3Dv2 and ScanNet, and RealEstate10K is not involved in the calibration process, demonstrating the cross-dataset generalization capability of our method. These results indicate that the performance gains stem from our geometry-aware quantization design rather than differences in the quantization regime. We note that QuantVGGT does not report real hardware latency for camera pose estimation on CO3Dv2, and its released implementation relies on pseudo-quantization. As latency is inherently tied to bit-width choices and hardware-specific optimizations, a strictly comparable latency evaluation is not available.

We further observe that generic PTQ methods such as GPTQ and AWQ suffer significant performance degradation under W4A16, highlighting the challenge of directly applying standard quantization techniques to VGGT. In contrast, QVGGT effectively preserves camera pose estimation accuracy across datasets, demonstrating both robustness and cross-domain generalization.

\subsection{Point Map Estimation}

Following CUT3R \citep{wang2025cut3r}, we evaluate the 3D reconstruction performance of QVGGT against state-of-the-art methods on the 7-Scenes \citep{shotton2013scene} and NRGBD \citep{azinovic2022neural} datasets. We adopt three widely used metrics: accuracy (Acc), completeness (Comp), and normal consistency (NC), where lower values indicate better performance for Acc and Comp, and higher values are preferred for NC.

Table \ref{tab:reconstruction_results} shows that QVGGT consistently achieves nearly lossless quantization compared to the FP16 baseline, while significantly outperforming generic quantization methods.

\begin{table}[t]
\large
\centering
\caption{Quantitative 3D reconstruction results on the 7-Scenes \citep{shotton2013scene} and NRGBD \citep{azinovic2022neural} datasets.}
\vspace{-1mm}
\label{tab:reconstruction_results}
\resizebox{\linewidth}{!}{%
\begin{tabular}{lcccccccc}
  \toprule
  \multirow{3}{*}{\textbf{Method}} & \multirow{3}{*}{\textbf{W/A}} & \multicolumn{6}{c}{\textbf{7-Scenes}} \\
  \cmidrule(lr){3-8}
  & & \multicolumn{2}{c}{Acc$\downarrow$} & \multicolumn{2}{c}{Comp$\downarrow$} & \multicolumn{2}{c}{NC$\uparrow$} \\
  \cmidrule(lr){3-4} \cmidrule(lr){5-6} \cmidrule(lr){7-8}
  & & Mean & Med. & Mean & Med. & Mean & Med. \\
  \midrule
  \rowcolor{gray!20}
  Baseline         & FP16       & 0.030 & 0.019 & 0.034 & 0.039 & 0.847 & 0.928 \\
  \midrule
  SmoothQuant \citep{xiao2023smoothquant}     & W8A8       & 0.067 & 0.060 & 0.058 & 0.042 & 0.702 & 0.760 \\
  GPTQ \citep{frantar2022gptq}         & W4A16  & 0.051 & 0.033 & \underline{0.053} & 0.025 & 0.802 & 0.880 \\
  AWQ \citep{lin2024awq}             & W4A16  & \underline{0.043} & \underline{0.023} & 0.054 & \underline{0.024} & \underline{0.819} & \underline{0.914} \\
  \rowcolor{lightblue}
  QVGGT (ours)         & W4A16  & \textbf{0.031} & \textbf{0.019} & \textbf{0.035} & \textbf{0.021} & \textbf{0.849} & \textbf{0.927} \\
  \midrule
  \multirow{3}{*}{\textbf{Method}} & \multirow{3}{*}{\textbf{W/A}} & \multicolumn{6}{c}{\textbf{NRGBD}} \\  %
  \cmidrule(lr){3-8}
  & & \multicolumn{2}{c}{Acc$\downarrow$} & \multicolumn{2}{c}{Comp$\downarrow$} & \multicolumn{2}{c}{NC$\uparrow$} \\
  \cmidrule(lr){3-4} \cmidrule(lr){5-6} \cmidrule(lr){7-8}
  & & Mean & Med. & Mean & Med. & Mean & Med. \\
  \midrule
  \rowcolor{gray!20}
  Baseline             & FP16       & 0.024 & 0.013 & 0.038 & 0.013 & 0.922 & 0.994 \\
  \midrule
  SmoothQuant \citep{xiao2023smoothquant}      & W8A8       & 0.062 & 0.052 & 0.066 & 0.054 & 0.769 & 0.778 \\ 
  GPTQ \citep{frantar2022gptq}        & W4A16  & 0.053 & 0.029 & \underline{0.051} & \underline{0.022} & 0.872 & 0.945 \\
  AWQ \citep{lin2024awq}             & W4A16  & \underline{0.047} & \underline{0.023} & 0.059 & 0.024 & \underline{0.891} & \underline{0.982} \\
  \rowcolor{lightblue}
  QVGGT (ours)         & W4A16  & \textbf{0.029} & \textbf{0.015} & \textbf{0.037} & \textbf{0.015} & \textbf{0.925} & \textbf{0.994} \\
  \bottomrule
\end{tabular}%
}
\end{table}

\subsection{Ablation Study}

\begin{table}[t]
  \small
  \centering
  \caption{Ablation study on CO3Dv2~\citep{reizenstein2021co3d} and NRGBD~\citep{azinovic2022neural}.
    \textbf{Q}: naively quantizing all blocks;
    \textbf{S}: selective mixed-precision quantization;
    \textbf{D}: token filtering with camera information compensation;
    \textbf{T}: task-aware scale-search mechanism.
    The \checkmark mark~means~the component is \textit{enabled}; --~means the component is~\textit{disabled}.}
    \vspace{-1mm}
  \label{tab:ablation}
  \begin{tabular}{cccccc}
    \toprule
    \multirow{2}{*}{\textbf{Q}} & \multirow{2}{*}{\textbf{S}} & \multirow{2}{*}{\textbf{D}} & \multirow{2}{*}{\textbf{T}} &
    \textbf{CO3Dv2} & \textbf{NRGBD} \\
    & & & & AUC@30$\uparrow$ & Accuracy Mean$\downarrow$ \\
    \midrule
    \rowcolor{gray!20}
    \checkmark & -- & -- & -- & 54.57 & 0.122 \\
    -- & \checkmark & -- & -- & 80.76 & 0.057 \\
    -- & \checkmark & \checkmark & -- & 85.91 & 0.054 \\
    \rowcolor{lightblue}
    -- & \checkmark & \checkmark & \checkmark & \textbf{89.39} & \textbf{0.029} \\
    \bottomrule
  \end{tabular}
\end{table}

\begin{table}[t]
\small
  \centering
  \caption{Ablation study of the 3D reconstruction performance with different calibration image numbers.}
  \vspace{-1mm}
  \label{tab:calibration_analysis}
  \begin{tabular}{ccc}
    \toprule
     \textbf{Calibration } & \textbf{CO3Dv2} & \textbf{NRGBD} \\
     \textbf{image numbers} & AUC@30$\uparrow$ & Accuracy mean$\downarrow$ \\
    \midrule
      16 & 87.86 & 0.035 \\
      32 & 89.17 & 0.032 \\
      128 & \textbf{89.39} & \textbf{0.029} \\
    \bottomrule
  \end{tabular}
  \vspace{-4mm}
\end{table}

\noindent\textbf{Quantization framework components.} We conduct ablation experiments on CO3Dv2 \citep{reizenstein2021co3d} and NRGBD \citep{azinovic2022neural} to assess the contribution of each component in QVGGT. The results are summarized in Table \ref{tab:ablation}.
Naively quantizing all blocks causes severe degradation on both benchmarks, confirming our earlier analysis that VGGT \citep{wang2025vggt} is highly sensitive to low-bit quantization, especially in blocks influencing the camera head. Selective mixed-precision quantization significantly improves performance on both datasets by preserving high-importance blocks in FP16 while quantizing others to 4 bits. This demonstrates that heterogeneous block sensitivity is a key obstacle for quantizing VGGT and that protecting critical layers is essential for retaining performance.
Next, applying token decoupling with camera information compensation reduces quantization-induced bias on the camera branch and yields improvements on both camera head and DPT \citep{ranftl21dpt} head predictions. Finally, adding the task-aware scale-search mechanism yields the best overall performance, achieving 89.39 (AUC@30) on CO3Dv2 and 0.029 (accuracy mean) on NRGBD.

\vspace{2mm}
\noindent\textbf{Robustness to calibration dataset size.} Table \ref{tab:calibration_analysis} evaluates QVGGT under different calibration set sizes. The results show that our method remains highly stable, with only minor performance variations when increasing the number of calibration images from 16 to 128. Although using 32 images yields a slight improvement, the overall trend indicates that QVGGT does not rely heavily on large calibration sets to obtain reliable activation statistics. This robustness highlights the practicality of QVGGT, as it achieves strong quantization performance even with a very small amount of calibration data.

\noindent\subsection{Efficiency Analysis}
We evaluate the efficiency of QVGGT in terms of peak memory consumption and inference latency.

\begin{figure}[t]
  \centering
  \includegraphics[width=0.98\linewidth]{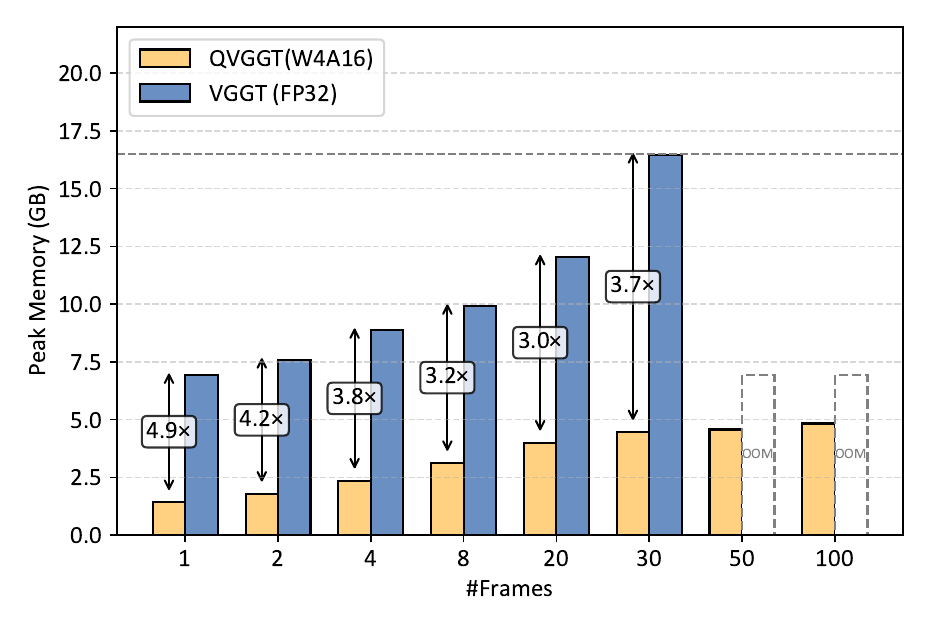}
  \vspace{-5mm}
  \caption{Peak memory reduction under different numbers of input frames (image resolution: 518$\times$518).}
  \label{fig:memory reduction}
\end{figure}

\vspace{2mm}
\noindent\textbf{Memory efficiency comparison.}
As shown in Figure \ref{fig:memory reduction}, our quantized model significantly reduces memory usage compared to the full-precision VGGT \citep{wang2025vggt}. With one input frame, QVGGT achieves a 4.9× reduction in peak memory, and this advantage remains consistent as the number of input frames increases. Even at 30 input frames, where full-precision VGGT nearly reaches the memory limit of a 24 GB GPU, QVGGT still maintains a 3.7× reduction in memory footprint. This demonstrates that our quantization framework not only compresses the model parameters but also effectively mitigates the activation memory overhead that scales with the number of frames.

\vspace{2mm}
\noindent\textbf{Latency efficiency comparison.}
Table~\ref{tab:opt} reports normalized memory and latency performance with two input frames. 
QVGGT achieves a 2.11$\times$ memory reduction and a 1.93$\times$ latency speedup relative to the FP16 baseline. 
These improvements indicate that low-bit weight quantization effectively reduces memory bandwidth pressure while improving computational efficiency during inference.

\vspace{-1mm}

\begin{table}[t]
  \small
  \centering
  \caption{Comparison among different methods in terms of the normalized memory and latency performance with 2 input frames.}
  \vspace{-1mm}
  \label{tab:opt}
  \begin{tabular}{lcc|c}  %
    \toprule
    \textbf{Method} & \textbf{W/A} & \textbf{Memory\,$\uparrow$} & \textbf{Latency\,$\downarrow$} \\
    \midrule
    \rowcolor{gray!20}
    Baseline        & FP16         & 1.00×                       & 1.00× \\ 
    QuantVGGT \citep{feng2025quantized}       & W8A8         & 1.93×                       & \textbf{2.17×} \\
    \rowcolor{lightblue}
    QVGGT (ours)    & W4A16        & \textbf{2.11× }   & 1.93× \\ 
    \bottomrule
  \end{tabular}
  \vspace{-4mm}
\end{table}

\section{Conclusion}

This work presents QVGGT, a post-training quantization framework tailored for VGGT. 
We analyze the heterogeneous quantization sensitivity of VGGT and the distinct behavior of camera tokens and register tokens, and address these challenges through a selective mixed-precision scheme, token filtering with camera information compensation, and a task-aware scale search guided by multi-head geometric consistency. 
Extensive experiments demonstrate that QVGGT achieves near-lossless 4-bit quantization, improving the deployment feasibility of high-fidelity 3D perception models. 
We note that QuantVGGT~\cite{feng2025quantized} is a concurrent work that also explores PTQ for VGGT. Our approach is complementary, focusing on geometry- and task-aware design, while QuantVGGT emphasizes distribution-oriented quantization strategies.

\section*{Acknowledgement}

This paper is supported by Young Scientists Fund of the National Natural Science Foundation of China (NSFC) (No. 62506305), Zhejiang Leading Innovative and Entrepreneur Team Introduction Program (No. 2024R01007), Key Research and Development Program of Zhejiang Province (No. 2025C01026), Scientific Research Project of Westlake University (No. WU2025WF003), Chinese Association for Artificial Intelligence (CAAI) \& Ant Group Research Fund - AGI Track (No. 2025CAAI-ANT-13). It is also supported by the research funds of the National Talent Program and Hangzhou Municipal Talent Program.

{
    \small
    \bibliographystyle{ieeenat_fullname}
    \bibliography{main}
}

\clearpage
\setcounter{page}{1}
\maketitlesupplementary

\setcounter{section}{0} %
\renewcommand{\thesection}{\Alph{section}} %

\section{Quantization Implementation Details}
\label{sec:supp1}
To facilitate reproducibility, we summarize the key hyperparameters and implementation details. Unless otherwise specified, we follow the default settings of the official VGGT~\citep{wang2025vggt} codebase.

\noindent\textbf{Weight quantization.}
We adopt 4-bit symmetric weight-only quantization with per-group scaling (group size = 128). To ensure optimal quantization accuracy while minimizing the computational cost of the calibration process, we randomly sample 128 calibration images from the Co3Dv2~\citep{reizenstein2021co3d} and ScanNet~\citep{dai2017scannet} datasets to collect activation statistics and perform scale search. 
These datasets provide complementary coverage of object-centric and indoor scene distributions, enabling more robust estimation of activation statistics for quantization. All linear layers in alternating-attention (AA) blocks and the camera head are quantized under these settings unless protected by selective mixed precision.

\noindent\textbf{Camera information compensation.}
For constructing the camera information compensation token, we perform PCA on the camera token activations collected from the calibration set and retain K = 32 principal components. The reconstructed camera information compensation token is normalized to match the average norm of patch tokens before being injected into the camera head during inference.

\noindent\textbf{Task-aware scale search.}
For the task-aware objective, we apply a balanced combination of head-specific losses. The depth head loss is weighted with coefficient $\alpha$ = 0.1, and the point-map head loss with coefficient $\beta$ = 0.1, while camera-head supervision and geometry-consistency loss are used with equal weight (weight=1.0).

\noindent\textbf{Computation cost of calibration.}
To assess the practicality of our PTQ pipeline, we further report the calibration overhead introduced by our framework. Using a calibration set of 128 images randomly sampled from CO3Dv2. The complete calibration process takes approximately 2 hours on a single RTX 4090 GPU. This overhead is modest compared to the full PTQ workflow and remains significantly lower than QAT-based approaches. These results demonstrate that our quantization framework is both computationally lightweight and suitable for deployment on widely accessible consumer hardware.

\section{Additional Analysis of Memory Efficiency}

VGGT~\citep{wang2025vggt} consists of 24 AA blocks, yet only 4 block outputs (the outputs of blocks 4, 11, 17, and 23) are consumed by downstream prediction heads.
However, the original VGGT implementation stores the activations of all 24 blocks during inference. When the number of input frames \( S \) increases, activation memory grows approximately as,
\[
\mathcal{O}(S \cdot H \cdot W \cdot C),
\]
where H and W are the height and width of the input image, respectively, this leads to prohibitive memory consumption when handling long sequences. To support long-sequence inference without overwhelming memory, we adopt a simplified method:
Intermediate outputs not used by any prediction head are discarded immediately; only the outputs of blocks 4, 11, 17, and 23 are stored during the inference.

\section{Additional Analysis of Camera Information Compensation}

\begin{table}[t]
\small
  \centering
  \renewcommand{\thetable}{A2}
  \caption{Ablation study of the 3D reconstruction performance with camera information compensation.}
  \vspace{-1mm}
  \label{tab:supp ablation}
  \begin{tabular}{ccc}
    \hline
     \textbf{Camera information} & \textbf{CO3Dv2} & \textbf{NRGBD} \\
     \textbf{compensation} & AUC@30$\uparrow$ & Accuracy mean$\downarrow$ \\
    \midrule
      -- & 84.51 & 0.029 \\
      \checkmark & \textbf{89.39} & 0.029 \\
    \bottomrule
  \end{tabular}
\end{table}

\begin{figure*}
    \centering
    \renewcommand{\thefigure}{A1}
    \includegraphics[width=1.0\textwidth]{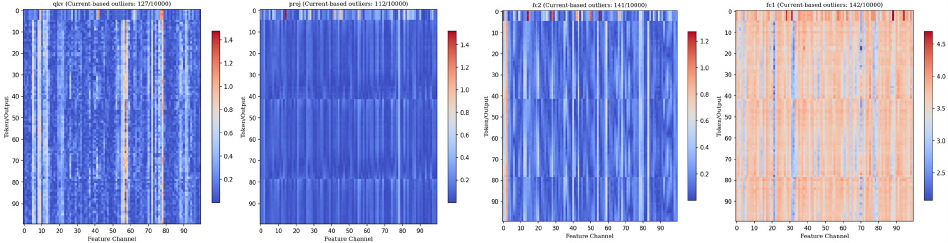}
    \caption{The outlier distribution heatmap of global block 22, global block 23, frame block 17, and frame block 19.}
\label{fig:supp2}
\end{figure*}

\begin{figure*}
    \centering
    \renewcommand{\thefigure}{A2}
    \includegraphics[width=1.0\textwidth]{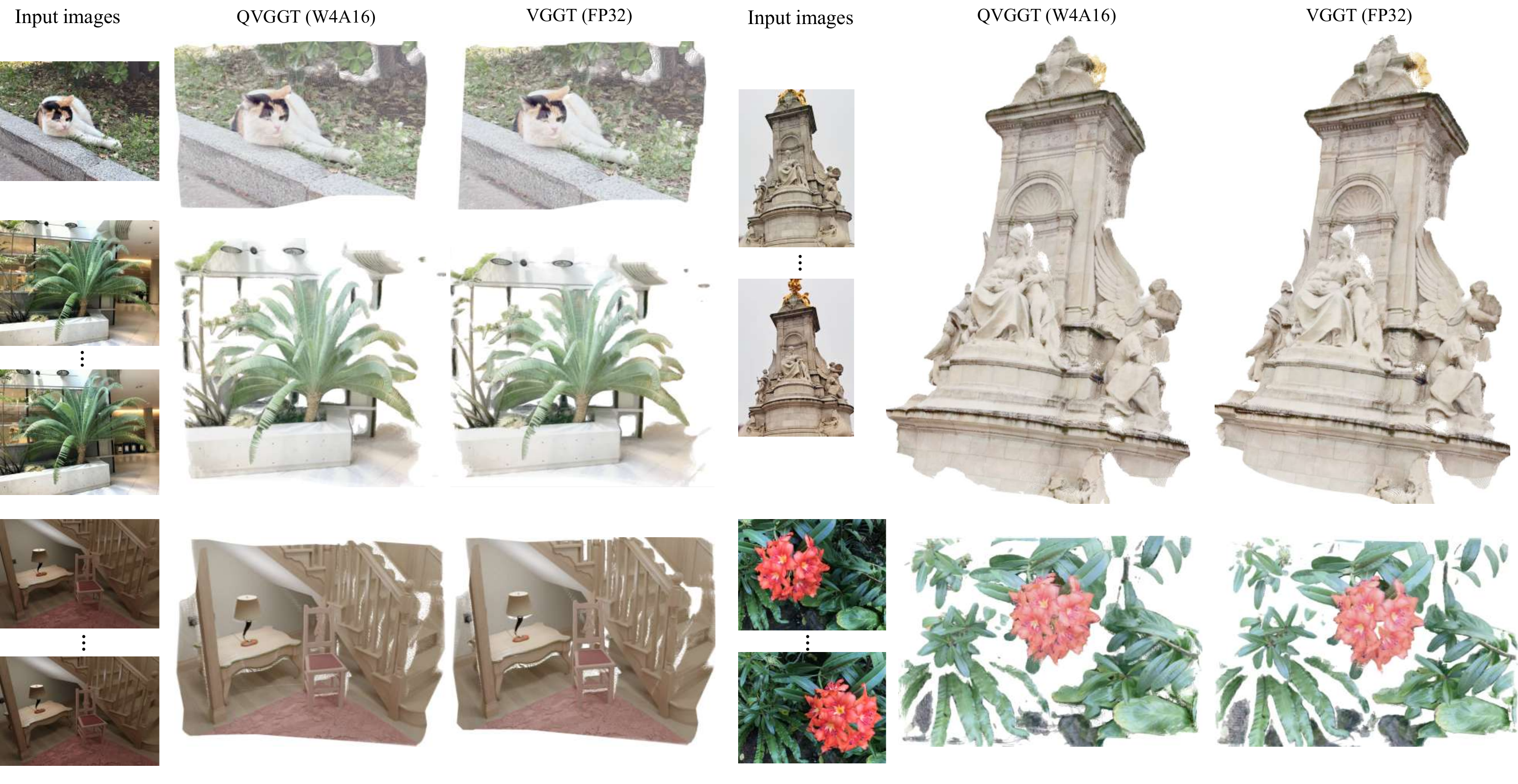}
    \caption{Comprehensive visualization comparisons of 3D reconstruction outputs between QVGGT and full-precision VGGT, covering representative scene categories: indoor and outdoor environments, real-world captured scenes, and synthetically generated scenes, as well as single-view and multi-view inference scenarios.}
\label{fig:supp1}
\end{figure*}

To further analyze the heterogeneous activation patterns that challenge low-bit quantization, we present additional 2D activation heatmaps across multiple blocks of VGGT (Fig. \ref{fig:supp2}). These visualizations consistently reveal a strong heavy-tailed distribution: camera tokens and register tokens exhibit substantially larger activation magnitudes and more irregular variance patterns compared to adjacent image tokens. This amplified amplitude persists across early, middle, and late transformer blocks, confirming that the outlier behavior of data-independent tokens is not an isolated case but a systematic property of VGGT.

These findings highlight two key implications. First, the pronounced deviation of camera/register tokens from the compact distribution of image tokens makes them disproportionately influential during activation-aware scale estimation, thereby causing unstable or suboptimal quantization scales. Second, since these tokens encode camera pose cues essential for camera pose prediction, simply excluding them from calibration reduces quantization noise but inevitably removes indispensable geometric information.

To explicitly verify this, we conduct an additional ablation study, presented in Table~\ref{tab:supp ablation}, which isolates the impact of our camera information compensation mechanism. Without compensation, the quantized model attains 84.51 AUC@30 on CO3Dv2~\citep{reizenstein2021co3d} and 0.054 mean error on NRGBD~\citep{azinovic2022neural}. With CIC enabled, the quantized model achieves 89.39 AUC@30 on CO3Dv2—representing a significant performance gain—while retaining the same depth accuracy as the uncompensated counterpart on NRGBD. This validates that our proposed compensation token successfully retrieves the missing geometric information, resolving the trade-off between stabilizing activation statistics during calibration and preserving critical camera pose cues.

\section{More Visualization Results}

We present more visual comparison results between full precision VGGT and QVGGT in Fig. \ref{fig:supp1}.

\end{document}